\documentclass[letterpaper, 10 pt, conference]{ieeeconf}  

\IEEEoverridecommandlockouts                              

\makeatletter
\def\input@path{{dotfiles-latex/}}
\makeatother
\usepackage{bais}

\usepackage{graphics} 
\usepackage{epsfig} 
\usepackage{mathptmx} 
\usepackage{times} 
\usepackage{amsmath} 
\usepackage{amssymb}  

\usepackage{tabularx}
\usepackage{booktabs}

\title{\LARGE \bf
  Identification of Unmodeled Objects from Symbolic Descriptions*
}

\author{Andrea Baisero, Stefan Otte, Peter Englert and Marc Toussaint
\thanks{*Work supported by the 3rdHand project, funded by the European Union
under the FP7 programme (FP7-ICT-2013-10610878).}
\thanks{All authors are with the Machine Learning and Robotics Lab, \newline
University of Stuttgart, Germany. \newline
{\tt\small <firstname.lastname>@ipvs.uni-stuttgart.de}}
}

\begin{document}

\maketitle
\thispagestyle{empty}
\pagestyle{empty}

\begin{abstract}

Successful human-robot cooperation hinges on each agent's ability to process
and exchange information about the shared environment and the task at hand.
Human communication is primarily based on symbolic abstractions of object
properties, rather than precise quantitative measures.  A comprehensive robotic
framework thus requires an integrated communication module which is able to
establish a link and convert between perceptual and abstract information.

The ability to interpret composite symbolic descriptions enables an autonomous
agent to
\begin{inlist}
\item operate in unstructured and cluttered environments, in tasks which
  involve unmodeled or never seen before objects; and
\item exploit the aggregation of multiple symbolic properties as an instance of
  ensemble learning, to improve identification performance even when the
  individual predicates encode generic information or are unprecisely grounded.
\end{inlist}

We propose a discriminative probabilistic model which interprets symbolic
descriptions to identify the referent object contextually w.r.t.\ the structure
of the environment and other objects.  The model is trained using a collected
dataset of identifications, and its performance is evaluated by quantitative
measures and a live demo developed on the PR2 robot platform, which integrates
elements of perception, object extraction, object identification and grasping.

\end{abstract}

\section{INTRODUCTION}

The human ability to compose and interpret object descriptions is a fundamental
one for the purpose of efficient collaboration during the concurrent
multi-agent execution of a complex task.  Humans are very skilled at guessing
games in which they have to identify objects given sparse, incomplete or even
mildly contradictory information.  Succeeding at such guessing games generally
requires two complementary skills: the ability to \emph{describe} an object
using a pre-specified language (\aka\ encoding the object identity), and the
ability to \emph{identify} an object given its description (\aka\ decoding the
object identity).  Both skills abstract the human ability to communicate about
objects in a wide range of environments.  As robotics research moves from
passive single-agent tasks to active manipulation in close collaboration with
humans, the ability to play such guessing games is bound to extend an
autonomous system's workspace.

\begin{figure}[t]
  \centering
  \resizebox{\columnwidth}{!}{%
    \begin{tikzpicture}
      \node (A) at (0,0)[draw=black]{%
        \includegraphics[width=\textwidth]{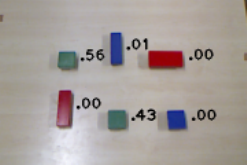}
      };
      \node[draw=black, below left=of A.south west, anchor=north] (B) {%
        \includegraphics[width=\textwidth]{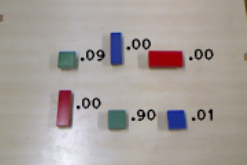}
      };
      \node[draw=black, below right=of A.south east, anchor=north] (C) {%
        \includegraphics[width=\textwidth]{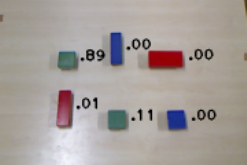}
      };
    \end{tikzpicture}
  }

  \caption{On the top, the identification distribution given an ambiguous
  description \desctt{green}. On the bottom left and right, the identification
distributions for more specific descriptions, respectively \desctt{green
bottom} and \desctt{green left}.}

  \label{fig:front_page}
\end{figure}

Humans share information and reason about objects using a symbolic language
consisting of low- and high-level (relational) qualitative properties, \eg\
\desctt{blue}, \desctt{laptop}, \desctt{next\_to}.  These symbolic predicates
can be flexibly composed into structured descriptions (\eg\ ``Get the
\desctt{thin} \desctt{book} \desctt{next\_to} the \desctt{white}
\desctt{laptop}'') which are typically tailored around the existing
environment.  The previous description might be suitable (or even redundant)
for some environments, but ambiguous in others, \eg\ in rooms where there are
many books and laptops.  The modularity of such descriptions enables them to
refer to an object which lacks a single prominent identifying quality, and to
exploit groups of uncertain properties which still contribute to the overall
identification process, in the style of ensemble models (\FIG{front_page}).

Concerning object recognition, a number of approaches are often used to
overcome the hurdle of perception.  Either the world is restricted to
a predefined structure and degree of sparsity; or all objects of interest are
known beforehand, with recognition systems tailored to previously acquired
object models; or fiducial markers are attached to the relevant objects; or the
objects can be easily distinguished by predefined attributes such as shape and
color.  Each of these approaches introduces a constraint on either the
workspace or the task, and thus on the operability of the autonomous system.

\section{RELATED WORK}\label{sec:related_work}

The language games of Steels \cite{steels1997constructing, steels1998origins,
steels2001language} consider the origin and use of language. Steels suggests
that the key to successful language grounding is to tightly couple it with
sensory-motor features and feedback. They represent the context without which
the language would be arbitrary and its effectiveness unverifiable.  A number
of \emph{language games}---dynamic and interactive multi-agent verbal
communication exercises---are proposed as a general framework to achieve this
goal. The \emph{guessing game}, in which agents have to draw each other's
attention to specific objects of the environment through verbal and non-verbal
communication, is the fundamental archetype for most language games.  For
example, \emph{talking heads} is a guessing game where the goal is to create
new terms and to converge to a common grounded language for object
identification.

Similar to Steels, we use a simple guessing game scenario as a setting to
develop the object identification model.  However, unlike in Steels' work where
the language symbols of each agent are learned and grounded independently, we
focus on the problem of how to map a given set of uncertain symbols to the
identity of an object in more complex scenes.  In our setting, we do not focus
on the question of how language arises, or how multiple agents can converge to
a similar language grounding; Rather, we assume that the language is already
shared, and explore in what way this language can be efficiently used to solve
the guessing game at hand.

Tellex \etal~\cite{tellex2014asking} develop an inverse semantics
approach to formulate recovery requests in a human understandable
format. Their focus is on how to transform the symbolic request to
a natural language one rather than generating the request at a
symbolic level.

A number of authors in the computer vision community have proposed a shift in
computer vision towards attribute detection, in contrast to the topic of
explicit object modeling.  Lampert~\etal~\cite{lampert2009learning} and
Farhadi~\etal~\cite{farhadi2009describing} demonstrate that appropriate visual
features can be extracted for this purpose and use standard machine learning
classifiers to learn attribute categories such as \emph{plastic}, \emph{round},
\emph{furry}, \etc.  We agree with the authors that an attribute-centric
framework for object representation can improve recognition and generalization
capabilities of a system.  However, non-goal-oriented attribute extraction by
itself does not serve any particular purpose apart from image indexing.  For an
autonomous system to make use of this type of object representation, the
extraction needs to be guided by a goal. We focus on descriptions for object
identification in cooperative multi-agent settings as a means to enhance
inter-agent communication about their environment.

Salvi~\etal~\cite{SalviMBS12} integrate language acquisition with affordance
models for actions and effects, thus grounding verbal task descriptions
together with perception and task execution representations;  In such a
setting, the language is learned through direct interaction with the
environment.  Their work however focuses on the description of tasks, rather
than the description of objects.

Schauerte~\etal~\cite{schauerte2014look} define a discriminative model for
object segmentation based on visual (through pointing) and verbal descriptions
of areas of interest.  They define a Conditional Random Field model which
integrates features of region contrast, pointing gestures and spoken
utterances.

In conclusion, the problem of computing optimal object identifications for a
given description which is contextual to the environment is a relatively novel
yet promising topic of interest.  We are not aware of work that integrate the
use of identification methods in a live demonstration on a physical robot.

\section{IDENTIFICATION MODEL}

Let $\Obj$ denote an arbitrary set of objects which may in principle exist and
which share a common set of measurable features;  in our setting, $\Obj$
contains all possible clusters of image pixels.  We define an
\emph{environment} $\env\subseteq\Obj$ as a finite non-empty subset of objects
which exist in a given context. A \emph{lexicon} $\lex$ is defined as a set of
symbolic labels which are known by all users, and a \emph{description}
$\desc\subseteq\lex$ as a subset of the lexicon, with the empty set being an
absolutely uninformative description, and the full set an overspecified and
almost surely highly contradictory description.

\subsection{Discriminative Identification Model}\label{sec:model}

We propose a model which generalizes standard multi-class Logistic Regression.
The identification task is also a labeling problem, although it differs from
multi-class classification with respect to a few key aspects.  Standard
classification is the problem of finding a mapping from an input vector
$\bm{x}\in\RRR^d$ to a label $y\in\{1,\ldots,m\}$ belonging to some predefined
set.  However, in our identification setting \begin{inlist} \item the number of
  classes and their assigned semantic are context dependent rather than fixed,
\item the output classes have features associated with them, and \item there is
no similar notion of an explicit input vector.\end{inlist}

The proposed discriminative model is derived from a joint log-linear parametric
form,
\begin{equation}
  \pr(\obj,\desc;\env) \propto \exp\phi(\obj,\desc;\env)\T\beta \;, \label{eq:model_a}
\end{equation}

\noindent where $\phi$ and $\beta$ are respectively the vector of
object-description features and the vector of model parameters.

We split the feature and parameter vectors $\phi$ and $\beta$ into independent
components (one for each symbol $\symb$ in the lexicon $\lex$):
\begin{align}
  \phi(\obj,\desc;\env) &= \bigtimes_{\symb\in\lex} \phi_\symb(\obj,\desc;\env) \;, \label{eq:split} \\
  \beta &= \bigtimes_{\symb\in\lex} \beta_\symb \;, \\ 
  \phi(\obj,\desc;\env)\T\beta &= \sum_{\symb\in\lex} \phi_\symb(\obj,\desc;\env)\T\beta_\symb \;.
\end{align}

We emphasize that this allows us to manually select different sets of features
$\phi_\symb$ to be associated with each symbol $\symb$.  This is relevant not
only because it allows us to provide prior knowledge---if available---into the
model, but also will plays out the role of indirect regularization during
training.


We further factorize the object-description features
$\phi_\symb(\obj,\desc;\env)$ into the product of an indicator
description-dependent symbol feature $\III[\symb\in\desc]$ and
description-independent object features $\phi_\symb(\obj;\env)$,
\begin{equation}
  \phi_\symb(\obj,\desc;\env) = \III[\symb\in\desc] \phi_\symb(\obj;\env),
\end{equation}

\noindent and use the indicator features to restrict the scope of the
summation, \begin{equation} \sum_{\symb\in\lex} \III[\symb\in\desc]
\phi_\symb(\obj;\env) = \sum_{\symb\in\desc} \phi_\symb(\obj;\env).
\end{equation}

Finally, the discriminative identification model is proportional to the joint
model for a fixed description $\desc$,
\begin{equation}
  \pr(\obj\cond\desc;\env) \propto \exp\sum_{\symb\in\desc} \phi_\symb(\obj;\env)\T\beta_\symb, \label{eq:itask_posterior}
\end{equation}

\noindent and its neg-log likelihood ($\nll$) is
\begin{align}
  \nll(\obj\cond\desc;\env) &= \log\sum_{\obj'\in\env}\exp\sum_{\symb\in\desc}\feat_\symb(\obj';\env)\T\beta_\symb \nonumber \\
                       &\pheq - \sum_{\symb\in\desc}\feat_\symb(\obj;\env)\T\beta_\symb.
\end{align}

\subsection{Training and Loss Function}





Multi-class Logistic Regression is usually trained using the neg-log likelihood
loss function.  That is an appropriate choice for typical classification
problems where training labels only indicate one single class as being the
correct one (\ie\ deterministic target distributions).  On the other hand, the
identification task has instances where the correct response is to exhibit
uncertainty through a non-deterministic posterior distribution (\eg\ in the
case of ambiguous or contradictory descriptions).  To account for this, we
train our model using the Kullback-Leibler loss function.  It is worth
mentioning that the Kullback-Leibler (KL) divergence represents a natural
generalization of the neg-log likelihood function, as they become equivalent
when the target distribution is deterministic.

Given a dataset $D = \{(\env_i, \desc_i, p_i)\}{}_i$ containing tuples of
environments $\env_i$, descriptions $\desc_i$ and target posterior
distributions $p_i$, the loss function is computed as
\begin{equation}
  L(\beta; D) = \sum_{(\env,\desc,p)\in D} \KL(p || q), \label{eq:loss}
\end{equation}

\noindent where $q$ is the model identity posterior distribution
\EQ{itask_posterior}.

The Jacobian $J$ and Hessian $H$ of \EQ{loss} are computed as
\begin{align}
  J &= \sum_{(\env,\desc,p)\in D} \Phi \left( \bm{q} - \bm{p} \right) \;,
  \label{eq:jacobian} \\
  H &= \sum_{(\env,\desc,p)\in D} \Phi \left[ \diag(q) - qq\T \right] \Phi\T
  \;. \label{eq:hessian}
\end{align}

\noindent where the $\Phi$ matrix aggregates the object-description feature
vectors $\phi(\obj,\desc;\env)$ column-wise.  Equations \EQ{jacobian} and
\EQ{hessian} allow the model to be trained using a variety of gradient-based
and Newton methods.  In our evaluation, we have used a standard implementation
of the BFGS optimization algorithm.

In this work, we enforce regularization implicitly by manually selecting
relevant features $\feat_\symb$ for each of the used symbols.  In the general
setting, where such expert knowledge may not be available, we expect a
normalizing term to be required to avoid overfitting.

\section{EVALUATION}\label{sec:evaluation}

We evaluate the proposed model on a domain consisting of wooden blocks of
different colors, shapes and sizes.  Model performance and generalization
properties are evaluated both using cross-validation on a collected data-set of
object identification tasks, and through a live demonstration on a PR2 robot
which integrates the identification model in a full working pipeline, from
perception to grasping.

\subsection{Lexicon and Features}

We use a lexicon $\Symb$ composed of location labels \desctt{left},
\desctt{right}, \desctt{top} and \desctt{bottom}; geometry labels
\desctt{thin}, \desctt{wide}, \desctt{short}, \desctt{tall}, \desctt{small} and
\desctt{big}; and chromatic labels \desctt{red}, \desctt{green}, \desctt{blue},
\desctt{yellow} and \desctt{white}.

For each object, the following features are computed from the respective
cluster of pixels in the image: \begin{inlist}\item average pixel position
  relative to the whole image resolution; \item cluster width and height
  relative to the whole image resolution; \item number of pixels relative to
the whole image size; \item average pixel hue; and \item pixel light
mode.\end{inlist}

As mentioned in \SECTION{model}, we can use \EQ{split} to guide the parameter
learning and enforce regularization by manually specifying which features
correlate with any given symbol.  In our work, we manually associate symbols
\desctt{left} and \desctt{right} with the horizontal position feature;
\desctt{top} and \desctt{bottom} with the vertical position feature;
\desctt{thin}, \desctt{wide}, \desctt{short} and \desctt{tall} with the shape
features; \desctt{small} and \desctt{big} with the size feature; \desctt{red},
\desctt{green}, \desctt{blue} and \desctt{yellow} with the hue chromatic
features; and \desctt{white} with the light chromatic feature.

\begin{figure*}
  \begin{subfigure}[b]{.196\textwidth}
    \centering
    \includegraphics[width=\textwidth]{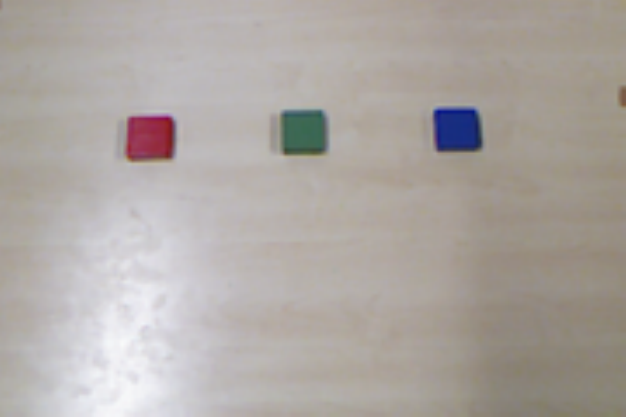}
    \caption{$\env_{1.1}$}
  \end{subfigure}
  \hfill
  \begin{subfigure}[b]{.196\textwidth}
    \centering
    \includegraphics[width=\textwidth]{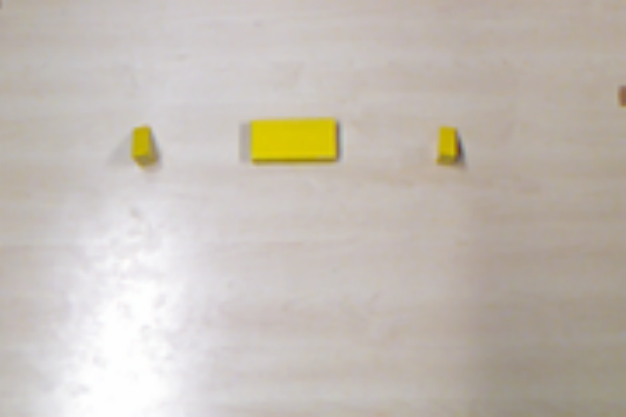}
    \caption{$\env_{2.4}$}
  \end{subfigure}
  \hfill
  \begin{subfigure}[b]{.196\textwidth}
    \centering
    \includegraphics[width=\textwidth]{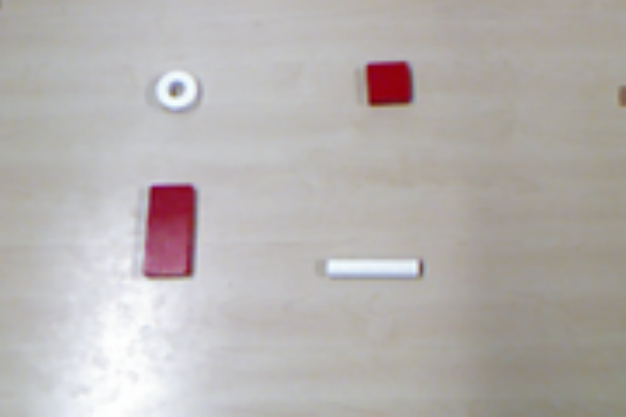}
    \caption{$\env_{3.5}$}
  \end{subfigure}
  \hfill
  \begin{subfigure}[b]{.196\textwidth}
    \centering
    \includegraphics[width=\textwidth]{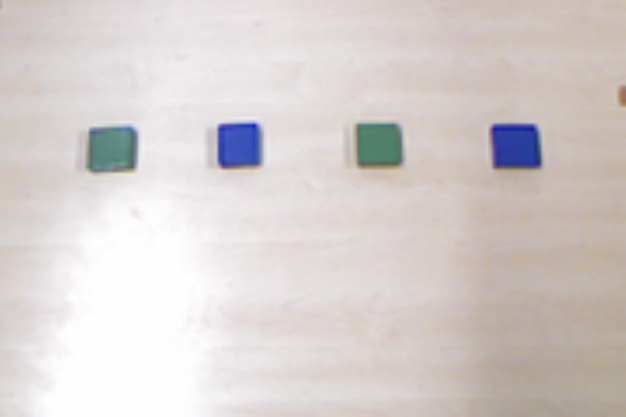}
    \caption{$\env_{4.2}$}
  \end{subfigure}
  \hfill
  \begin{subfigure}[b]{.196\textwidth}
    \centering
    \includegraphics[width=\textwidth]{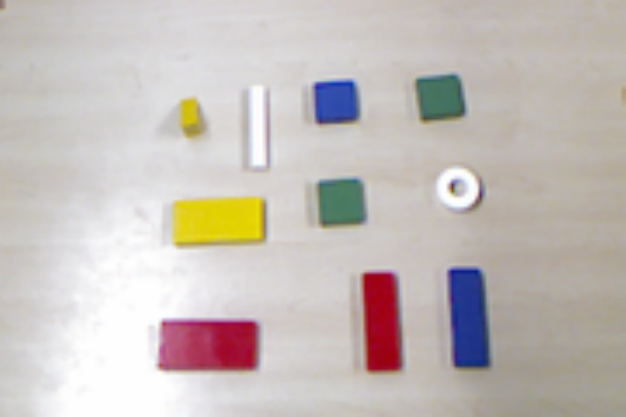}
    \caption{$\env_{5.1}$}
  \end{subfigure}
  \\

  \begin{subfigure}[b]{.196\textwidth}
    \centering
    \includegraphics[width=\textwidth]{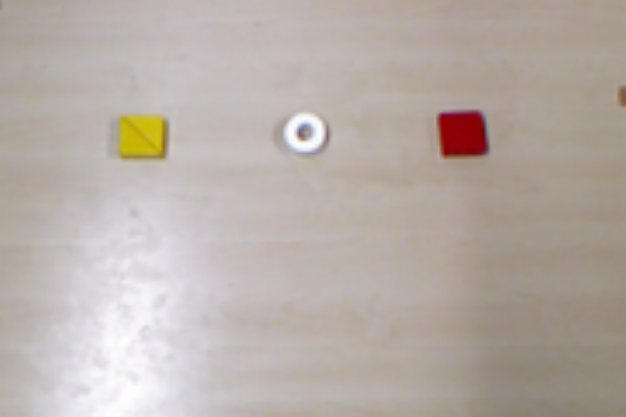}
    \caption{$\env_{1.4}$}
  \end{subfigure}
  \hfill
  \begin{subfigure}[b]{.196\textwidth}
    \centering
    \includegraphics[width=\textwidth]{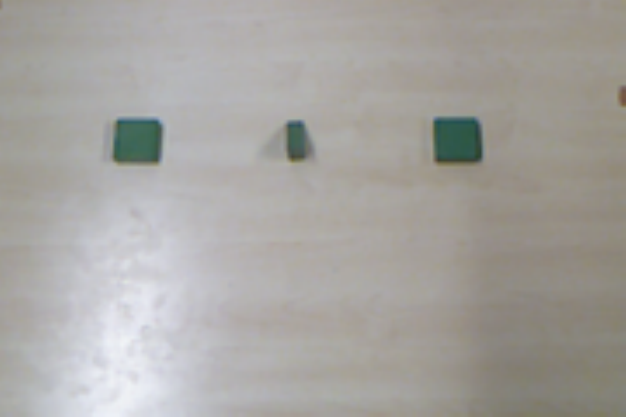}
    \caption{$\env_{2.2}$}
  \end{subfigure}
  \hfill
  \begin{subfigure}[b]{.196\textwidth}
    \centering
    \includegraphics[width=\textwidth]{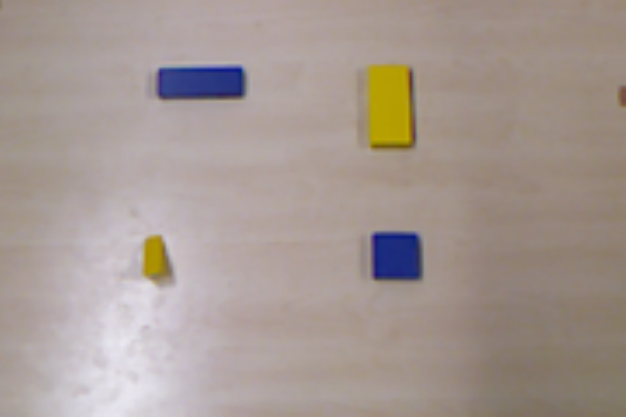}
    \caption{$\env_{3.3}$}
  \end{subfigure}
  \hfill
  \begin{subfigure}[b]{.196\textwidth}
    \centering
    \includegraphics[width=\textwidth]{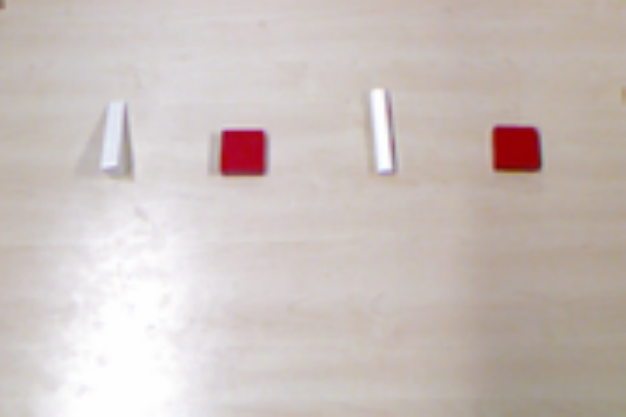}
    \caption{$\env_{4.5}$}
  \end{subfigure}
  \hfill
  \begin{subfigure}[b]{.196\textwidth}
    \centering
    \includegraphics[width=\textwidth]{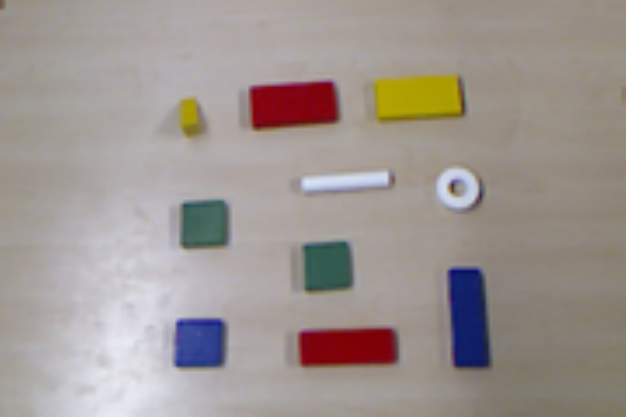}
    \caption{$\env_{5.2}$}
  \end{subfigure}
  \\

  \caption{The images depict 10 of the 22 environments used for training.  Each
    column contains 2 samples from each category.  The first category contains
    environments where three differently colored blocks are positioned side by
    side;  the second category contains environments in which blocks of the
    same color have to be identified through their geometry of position.  The
    third category contains blocks of 2 colors positioned in way that each
    block has a variery of correct descriptions which may be used to identify
    it.  The fourth category contains environments in which a correct
    identification can sometimes only be achieved by learning an appropriate
    trade-off between the relevancy of positional and chromatic features. The
    fifth category only contains 2 environments in which the many objects are
    located in no particular order and without any pattern.}

  \label{fig:dtask_data}
\end{figure*}

\subsection{Training Data}

To train and evaluate the model, we collected a data-set of object
descriptions and one of object identifications.

\subsubsection{Description Data}

The data-set consists of 22 environments, which we denote as
$\env_{[1\textnormal{-}4].[1\textnormal{-}5]}$ and
$\env_{5.[1\textnormal{-}2]}$ and differ in the number of objects, their
properties, and/or their disposition on the table (\FIG{dtask_data}).  We
partition these into 5 categories $\gamma_i = \{\env_{i.*}\}$ which broadly
share some underlying theme or pattern, respectively containing 5, 5, 5, 5 and
2 environments.  Multiple descriptions are provided for each object in each
environment, ranging from overly-specific to ambiguous ones, for a total of 660
descriptions in the whole dataset.

\subsubsection{Identification Data}

Each description is interpreted by 10 users, resulting in 6600 identification
data-points.
The users are instructed, for each description, to select all
objects which are plausible subjects of the description according to their
interpretation. The corresponding target distribution is constructed such that
the objects which have not been selected assume a low probability mass of
0.005, while the rest of the mass is uniformly distributed among all selected
objects.

\subsection{Performance Statistics}

Given an environment $\env$, we define a partition of the dataset $D = D_\env
\bigcup D_{\setminus\env}$, where $D_\env$ contains all data regarding
environment $\env$, and $D_{\setminus\env}$ is its complement.  Given a
category $\gamma$, we further define a partition of the dataset $D = D_\gamma
\bigcup D_{\setminus\gamma}$, where $D_\gamma = \bigcup_{\env\in\gamma}
D_\env$, and $D_{\setminus\gamma}$ is the complement.

We perform cross-validation using both environment-induced partitions and
category-induced partitions.  Performing cross-validation on the
category-induced partitions further ensures that the model evaluation does not
suffer from overfitting, due to the shared high-level patterns which however
don't influence the identification task.

In the first evaluation, we iterate through all environments $\env$ and use
$D_{\setminus\env}$ to train the model and $D_\env$ to evaluate it.  We then
repeat the process iterating through al categories $\gamma$, using
$D_{\setminus\gamma}$ and $D_\gamma$ respectively for training and testing.

The results---which are summarized in \TABLE{cv}---indicate that the model
suffers the most in situations where chromatic features are undiscriminative
and irrelevant.  The lower-than-average performance on category $\gamma_5$ may
be a consequence of the number of objects, which is at least double compared to
all the other environments in the data-set.

\begin{table}
  \centering

  \caption{Evaluation Statistics computed by cross-validation using
    environment-based splits and category-based splits.  Column t\_lklh
    contains the fraction of posterior identity mass $\pr(\obj\cond\desc)$
    which is correctly assigned to the objects selected by the target
  distribution;  this is approximatively $q\T p$, with $q$ and $p$ from
\EQ{loss}. Column $\KL$ contains the average KL-divergence loss per
identification task [nats].}

  \begin{tabularx}{\columnwidth}{r|cc|r|cc|r|cc}
    & t\_lklh & $\KL$ & & t\_lklh & $\KL$ & & t\_lklh & $\KL$ \\
    \toprule
$\env_{1.1}$ & 95.2\% & 0.04 & $\env_{3.1}$ & 92.0\% & 0.12 & $\env_{5.1}$ & 82.0\% & 0.29 \\
$\env_{1.2}$ & 95.0\% & 0.04 & $\env_{3.2}$ & 91.5\% & 0.14 & $\env_{5.2}$ & 78.4\% & 0.32 \\
$\env_{1.3}$ & 93.5\% & 0.05 & $\env_{3.3}$ & 90.3\% & 0.18 \\
$\env_{1.4}$ & 94.9\% & 0.04 & $\env_{3.4}$ & 88.9\% & 0.21 & $\gamma_1$ & 94.9\% & 0.04 \\
$\env_{1.5}$ & 96.0\% & 0.04 & $\env_{3.5}$ & 89.8\% & 0.18 & $\gamma_2$ & 85.5\% & 0.24 \\
$\env_{2.1}$ & 76.4\% & 0.39 & $\env_{4.1}$ & 92.7\% & 0.11 & $\gamma_3$ & 90.3\% & 0.17 \\
$\env_{2.2}$ & 88.4\% & 0.16 & $\env_{4.2}$ & 92.4\% & 0.16 & $\gamma_4$ & 91.5\% & 0.13 \\
$\env_{2.3}$ & 76.5\% & 0.36 & $\env_{4.3}$ & 91.6\% & 0.11 & $\gamma_5$ & 80.5\% & 0.31 \\
$\env_{2.4}$ & 93.5\% & 0.08 & $\env_{4.4}$ & 88.9\% & 0.19 \\
$\env_{2.5}$ & 89.2\% & 0.23 & $\env_{4.5}$ & 92.0\% & 0.11 & $\avg$ & 86.3\% & 0.22 \\
    \bottomrule
  \end{tabularx}

  \label{tab:cv}
\end{table}

\subsection{Online Real-World Demonstration}

We further illustrate the identification model in an integrated real-world
demonstration in which a PR2 robot observes blocks arbitrarily disposed on a
table and, upon receiving the description of one of the blocks, identifies and
grasps it.

\begin{figure*}
  \begin{subfigure}[b]{.246\textwidth}
    \centering
    \includegraphics[width=\textwidth]{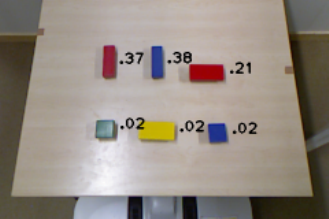}
    \caption{\bfseries \desctt{top}}
  \end{subfigure}
  \hfill
  \begin{subfigure}[b]{.246\textwidth}
    \centering
    \includegraphics[width=\textwidth]{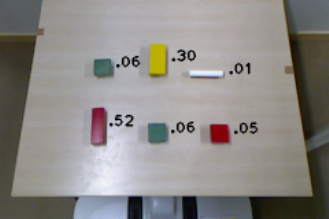}
    \caption{\bfseries \desctt{tall}}
  \end{subfigure}
  \hfill
  \begin{subfigure}[b]{.246\textwidth}
    \centering
    \includegraphics[width=\textwidth]{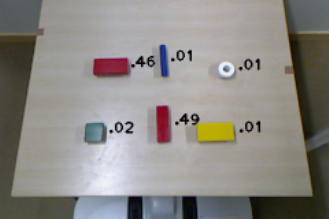}
    \caption{\bfseries \desctt{red}}
  \end{subfigure}
  \hfill
  \begin{subfigure}[b]{.246\textwidth}
    \centering
    \includegraphics[width=\textwidth]{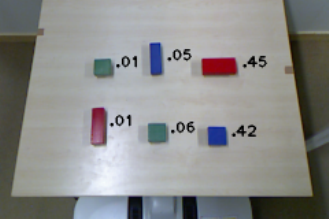}
    \caption{\bfseries \desctt{right}}
  \end{subfigure}
  \\

  \begin{subfigure}[b]{.246\textwidth}
    \centering
    \includegraphics[width=\textwidth]{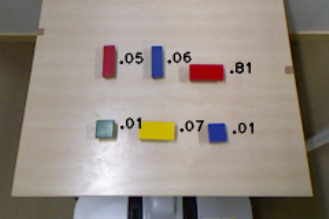}
    \caption{\bfseries \desctt{top wide}}
  \end{subfigure}
  \hfill
  \begin{subfigure}[b]{.246\textwidth}
    \centering
    \includegraphics[width=\textwidth]{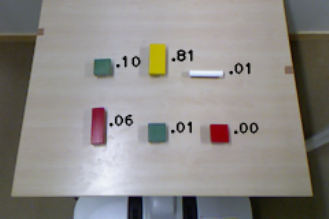}
    \caption{\bfseries \desctt{tall top}}
  \end{subfigure}
  \hfill
  \begin{subfigure}[b]{.246\textwidth}
    \centering
    \includegraphics[width=\textwidth]{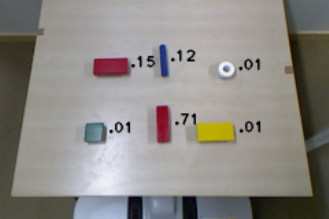}
    \caption{\bfseries \desctt{red thin}}
  \end{subfigure}
  \hfill
  \begin{subfigure}[b]{.246\textwidth}
    \centering
    \includegraphics[width=\textwidth]{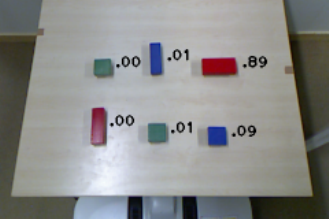}
    \caption{\bfseries \desctt{right wide}}
  \end{subfigure}
  \\

  \begin{subfigure}[b]{.246\textwidth}
    \centering
    \includegraphics[width=\textwidth]{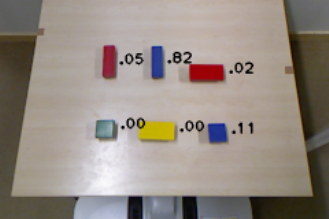}
    \caption{\bfseries \desctt{top blue}}
  \end{subfigure}
  \hfill
  \begin{subfigure}[b]{.246\textwidth}
    \centering
    \includegraphics[width=\textwidth]{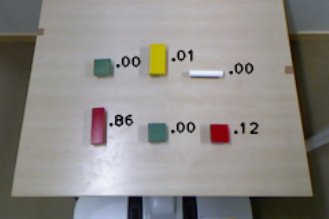}
    \caption{\bfseries \desctt{tall red}}
  \end{subfigure}
  \hfill
  \begin{subfigure}[b]{.246\textwidth}
    \centering
    \includegraphics[width=\textwidth]{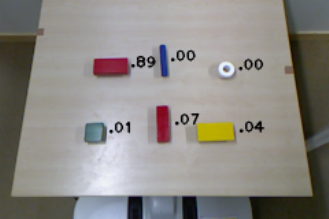}
    \caption{\bfseries \desctt{red wide}}
  \end{subfigure}
  \hfill
  \begin{subfigure}[b]{.246\textwidth}
    \centering
    \includegraphics[width=\textwidth]{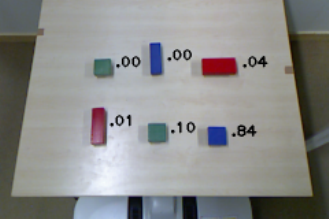}
    \caption{\bfseries \desctt{right bottom}}
  \end{subfigure}
  \\

  \caption{Object identity posterior distributions for a selection of
    previously unseen environments.   Each column represents the same
    environment where different descriptions are provided.  The first row
    depicts the model output when provided with inherently ambiguous
    descriptions.  In the second and third rows, the descriptions are extended,
    thus resolving the ambiguity in one way or another. Each environment is
    novel and demonstrates the model's generalization properties in the sense
  that, while the same objects were used during training, these particular
dispositions have never been seen before.}

  \label{fig:test}
\end{figure*}

The domain of this integrated demonstration is similar to the one in the
training data-set, but allows us to test performance in a wider range of
environments which differ even more from the ones used for training.
\FIG{test} illustrates a selection of results obtained during the execution
of the integrated demonstrations.

\begin{figure*}
  \resizebox{\textwidth}{!}{\input{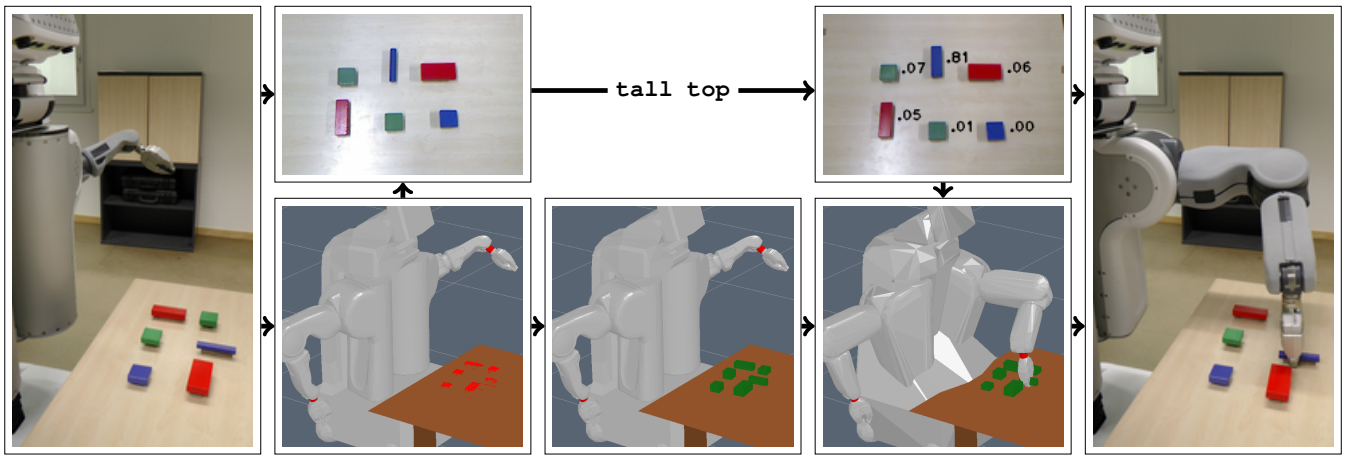}\unskip}

  \caption{Object grasping demonstration on a PR2.  The Object Recognition
    Kitchen ors package provides table-top plane extraction and object clusters
    (in red).  Box models are fitted (in green) in order to find an adequate
    grasping direction, while our model computes the identity distribution upon
  receiving the description \desctt{tall top}.  Finally, the robot proceeds to
successfully grasp the correct object along an appropriate axis.}

  \label{fig:demo}
\end{figure*}

The demonstration pipeline depicted in \FIG{demo} mainly consists of 4
components: an object extration module, the identification model presented in
this work, a grasping heuristic valid for our blocks domain, and a trajectory
optimization routine for grasping.

\subsubsection{Object Extraction}

We use the Object Recognition Kitchen (ORK) \cite{ork_ros}, a ROS package which
specializes in plane extraction and (modeled) object recognition, but which
also clusters all points which do not match any of the extracted plane.

We select all clusters in 3D space which appear above the table plane, project
them onto the 2D image and compute their convex-hulls, which represent binary
masks through which the previously mentioned image features can be computed for
each object.

\subsubsection{Grasping Heuristic}

We heuristically determine the optimal grasping position for an object
by finding the horizontal direction vector $d$ along which the projected object
point-cloud $P$ assumes the minimal thickness:
\begin{align*}
  \argmin_d & \left[ \max_{x\in P} d\T x - \min_{x\in P} d\T x \right] \\
            & \text{s.t.} \\
      \|d\| &= 1, \\
      d\T z &= 0,
\end{align*}

\noindent where $z$ is a vertical vector in the world frame.  Grasping position
is set as the mean position of the point-cloud $P$.

\subsubsection{Trajectory Optimization}

We use k-order Markov motion optimization \cite{14-toussaint-KOMO} to plan the
motion for grasping the object.  We define a cost function that consists of the
gripper position and orientation during the grasp.  We additionally ensure safe
motions by including collision avoidance and joint limit limit constraints into
the problem formulation.

\section{CONCLUSIONS}

In this work we propose and evaluate a discriminative model which is able to
interpret object descriptions and decode the intended object identity.
Quantitative and qualitative tests demonstrate positive identification
capabilities, and an implementation on a PR2 robot demonstrates the usage in a
real-world scenario in which unmodeled objects are being recognized by their
description.  


\subsection{Further Work}

The work presented in this submisison is not meant to be conclusive, but rather
represents a stepping stone towards more sophisticated methods to bridge the
gap between perception and geometric and symbolic representations.  We identify
the following extensions and topics of interest for the further development.

While the focus of this work has been on the topic of object identification,
the whole premise of successful human-robot communication requires a
bidirectional exchange of information.  An extension of direct interest
consists in building on top of the discriminative identification model to
obtain a generative description model which is able to produce sparse and
minimal symbolic descriptions understandable by humans.

Another extension of immediate interest involves the usage of relational
symbols.  Their inclusion in the framework would greatly extend the space of
existing descriptions in an environment, which produces two benefits: it
increases the chance that an appropriate description exists for any given
object---albeit this is true for any extension of the lexicon--- and it
potentially decreases the minimal complexity of a description required to
describe an object.

The experiments presented in our work have considered a relatively simple
lexicon, and simple object features extracted exclusively from image segments.
Future effort should also focus on applying the model in more extensive domains
where a bigger lexicon implies a wider range of possible descriptions, and in
which more informative features (\eg 3D point-cloud features) may be required
to successfully perform the identification task.

\bibliographystyle{abbrv}
\bibliography{refs}

\end{document}